\documentclass[11pt]{article}
\usepackage[final]{acl}

\usepackage{times}
\usepackage{latexsym}
\usepackage{amsmath} 
\usepackage{amssymb}

\usepackage[most]{tcolorbox}
\usepackage[utf8]{inputenc}   
\usepackage[T1]{fontenc}
\usepackage{listings}
\usepackage{listingsutf8}
\usepackage{array}
\usepackage{float}
\usepackage{dashrule}
\usepackage{microtype}

\usepackage{inconsolata}

\usepackage{graphicx}

\title{Who Judges the Judge? Evaluating LLM-as-a-Judge for French Medical open-ended QA}

\author{
  \textbf{Ikram Belmadani}\textsuperscript{1,2*}\quad
  \textbf{Oumaima El Khettari}\textsuperscript{2*}\quad \\
  \textbf{Pacôme Constant dit Beaufils}\textsuperscript{3,4} 
  \textbf{Richard Dufour}\textsuperscript{2}\quad
  \textbf{Benoit Favre}\textsuperscript{1,5} \\ \\\\
  \textsuperscript{1} Aix-Marseille Univ., CNRS, LIS UMR 7020, 13000 Marseille, France \\
  \textsuperscript{2} Nantes Univ., École Centrale Nantes, CNRS, LS2N, UMR 6004, 44000 Nantes, France \\
  \textsuperscript{3} Nantes Université, CHU Nantes, PHU 11: Santé Publique, \\ Clinique des données, INSERM, CIC 1413, 44000 Nantes, France \\
  \textsuperscript{4} Nantes Université, CNRS, INSERM, L'institut du thorax, 44000 Nantes, France \\
  \textsuperscript{5} Université Grenoble Alpes, CNRS, INRIA, Grenoble INP, \\
  LIG UMR 5217, 38100 Grenoble, France \\
  \textbf{Correspondence:} \href{mailto:email@domain}{\texttt{\{first.last\}@\{univ-amu.fr, univ-nantes.fr, chu-nantes.fr\}}}
}

\begin{document}

\maketitle
\renewcommand{\thefootnote}{\fnsymbol{footnote}}
\footnotetext[1]{Equal contribution.}
\def\thefootnote{\arabic{footnote}}
\begin{abstract}
Automatic evaluation of medical open-ended question answering (OEQA) remains challenging due to the need for expert annotations. We evaluate whether large language models (LLMs) can act as judges of semantic equivalence in French medical OEQA, comparing closed-access, general-purpose, and biomedical domain-adapted models. Our results show that LLM-based judgments are strongly influenced by the model that generated the answer, with agreement varying substantially across generators. Domain-adapted and large general-purpose models achieve the highest alignment with expert annotations. We further show that lightweight adaptation of a compact model using supervised fine-tuning (SFT) and Group Relative Policy Optimization (GRPO) substantially improves performance and reduces generator sensitivity, even with limited data. Overall, our findings highlight the need for generator-aware evaluation and suggest that carefully adapted small models can support scalable evaluation in low-resource medical settings.
\end{abstract}
\begin{figure*}[htbp]
\centering
\includegraphics[width=\textwidth]{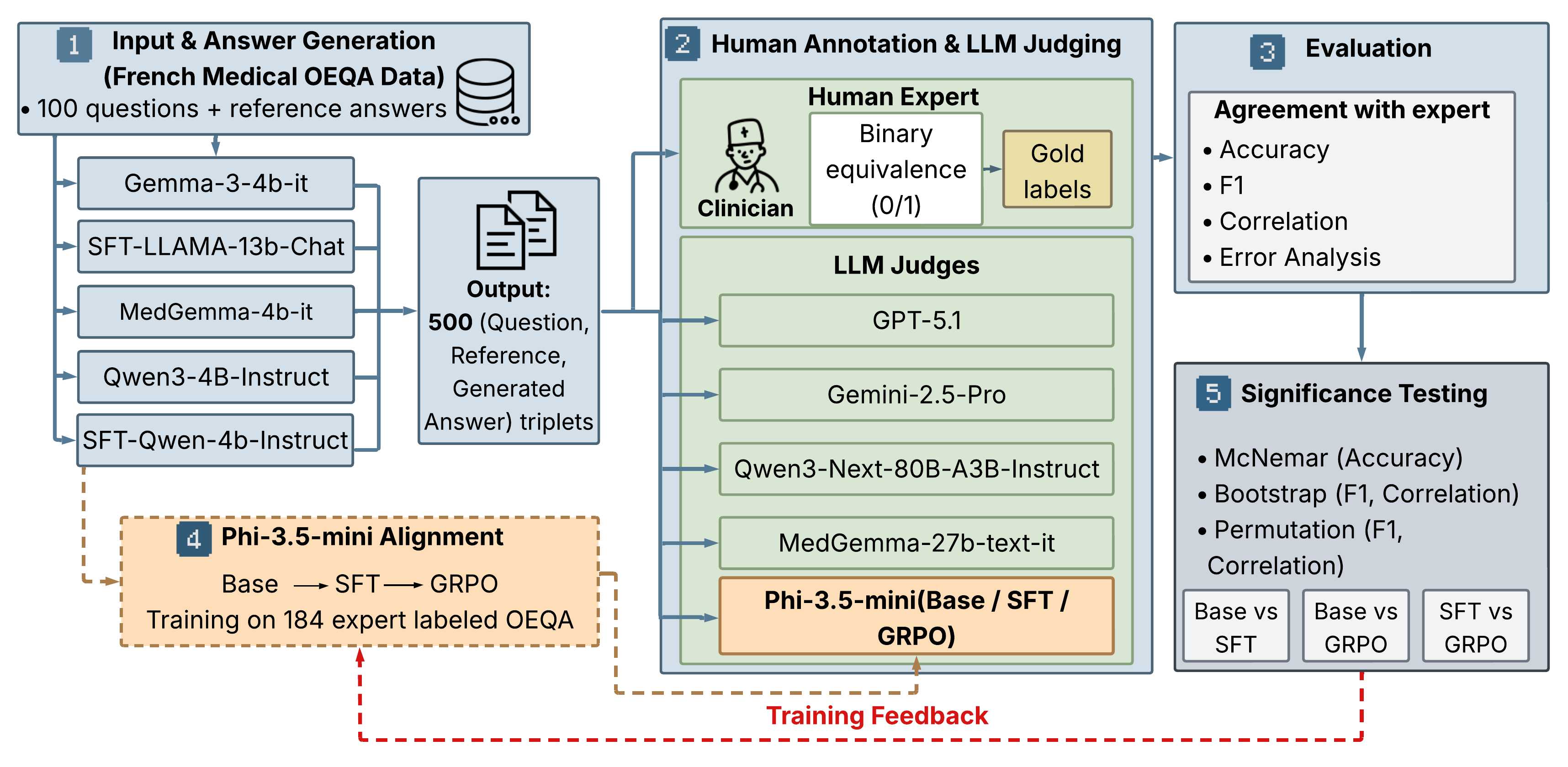}
\vspace{-1.5em}
\caption{Overview of the evaluation and alignment pipeline. French medical OEQA questions are answered by multiple LLM generators and annotated by a clinician for binary semantic equivalence. LLM evaluators judge the same instances using an identical prompt. A lightweight evaluator (Phi-3.5-mini) is further aligned via SFT and GRPO, and improvements are validated through paired significance testing.}
\label{overview}
\end{figure*}

\section{Introduction}

Automatically evaluating OEQA systems in specialized domains such as medicine remains a major challenge. Correctness depends on semantic fidelity, factual accuracy, and clinical relevance rather than surface similarity, making automatic metrics such as BLEU~\citep{papineni-etal-2002-bleu}, ROUGE~\citep{lin-2004-rouge}, or BERTScore~\citep{zhang2019bertscore} poorly suited for capturing medically valid paraphrases or detecting subtle but clinically meaningful errors. As a consequence, progress in medical LLM development critically depends on reliable evaluation. Yet, most current evaluation pipelines rely on multiple-choice QA (MCQA) benchmarks such as PubMedQA~\citep{jin-etal-2019-pubmedqa}, MedQA~\citep{jin2021disease}, and MedMCQA~\citep{pmlr-v174-pal22a}. These benchmarks evaluate classification performance, where the model selects the correct option from a predefined set, rather than generating a clinically valid free-form answer. As a result, strong MCQA performance does not necessarily reflect a model’s ability to perform open-ended medical reasoning. OEQA provides a more informative and clinically realistic evaluation setting, but it is also expensive, since it requires domain experts performing manual evaluation.

The emerging paradigm of \emph{LLM-as-a-Judge} offers a potential alternative to human expert evaluation, with LLMs showing encouraging performance in general-domain evaluation tasks~\citep{gu2025surveyllmasajudge}. However, several questions remain open in the context of medical OEQA: their reliability in specialized medical settings, their sensitivity to the model producing the answer, and their applicability to OEQA, a setting that remains largely unexplored compared to MCQA-based evaluation pipelines. Furthermore, it is unclear whether compact models can be effectively aligned to serve as trustworthy evaluators with only limited expert supervision, enabling more accessible and cost-efficient evaluation workflows.

This work presents a preliminary investigation of these questions. Although our dataset is small, the study constitutes a first step toward scalable, domain-aware evaluation frameworks for French medical OEQA. Focusing on French is particularly important, as it provides a testbed for LLM-as-a-Judge evaluation beyond English and enables analysis in a medical setting that differs from US/UK clinical practice and terminology. These characteristics make French medical OEQA a meaningful case for studying the robustness and transferability of LLM-based evaluation methods and for informing larger-scale future efforts.

We structure our study around the following research questions:

\begin{itemize}
    \item \textbf{RQ1:} How accurately do current LLMs reproduce expert judgments of semantic equivalence in French medical OEQA?
    \item \textbf{RQ2:} How sensitive are LLM-based evaluators to the specific answer generators they assess?
    \item \textbf{RQ3:} Can small models be effectively aligned to act as domain-aware medical judges using limited expert supervision?
    \item \textbf{RQ4:} Does reinforcement optimization on limited training data improve evaluator reliability beyond supervised fine-tuning on the same equivalence task?
\end{itemize}

This study offers three main contributions:

\begin{enumerate}
    \item A systematic assessment of LLM-as-a-Judge in French medical OEQA, showing how different families of evaluators, closed-access, open-source, and biomedical models, align with expert semantic equivalence judgments.
    \item An empirical analysis revealing that LLM judges are not generator-invariant, highlighting consistent evaluation biases linked to answer style, model family, and domain adaptation, and underscoring the need for generator-aware evaluation design.
    \item A demonstration that compact models can be effectively adapted into reliable evaluators, where lightweight SFT and GRPO substantially improve judgment stability and alignment with medical experts, suggesting a viable path for low-resource, domain-specific evaluation.
\end{enumerate}

All data and code used in this study are released publicly: \url{https://github.com/ikram28/LLM-as-a-Judge-for-medical-oeqa}.

\section{Related Work}
\subsection{Automatic Evaluation of OEQA}

Automatic evaluation of OEQA has traditionally relied on lexical or embedding-based metrics such as BLEU~\citep{papineni-etal-2002-bleu}, ROUGE~\citep{lin-2004-rouge}, METEOR~\citep{banerjee-lavie-2005-meteor}, and BERTScore~\citep{zhang2019bertscore}. Although these metrics capture surface-level similarity, they are insensitive to paraphrastic variation and often fail to reflect factual correctness or reasoning quality, resulting in weak correlation with human judgments in generative QA settings~\citep{liu-etal-2023-g}. These limitations are amplified in biomedical contexts, where clinically appropriate answers may diverge from reference formulations and correctness depends on precise domain knowledge~\citep{yim2025morqa}. Medical QA thus presents unique evaluation challenges, involving specialized terminology and subtle semantic distinctions central to clinical validity~\citep{berger2025reasoningllmsmedicaldomain}. Despite recent progress in French biomedical NLP, evaluation practices remain underdeveloped, with most approaches still relying on lexical overlap, limiting the reliability of current assessments~\citep{yim2025morqa}.

\subsection{LLM-as-a-Judge}
The use of LLMs as automated evaluators, commonly referred to as \textit{LLM-as-a-Judge}, has gained significant momentum as a scalable alternative to manual assessment. A comprehensive survey~\citep{gu2025surveyllmasajudge} highlights this paradigm’s promise for achieving consistent and cost-effective assessment, while also underscoring the challenges of reliability and bias control. Early work showed that GPT-4 can approximate human judgments in tasks such as summarization and dialogue when guided by structured prompting~\citep{liu-etal-2023-g}. Subsequent studies demonstrated that LLM judges can outperform traditional metrics in semantic evaluation tasks such as automated query parsing~\citep{10.3389/fdata.2025.1611389} and can be further stabilized through preference-based calibration~\citep{liu-etal-2024-calibrating}. Practical toolkits like ~\citet{li-sun-2025-easyjudge} have begun standardizing these methodologies for multi-criteria evaluation.

Despite this progress, growing evidence highlights systematic biases in LLM-based evaluators, such as a tendency to favor fluent or verbose responses even when they are factually incorrect, as well as sensitivity to stylistic or demographic cues unrelated to answer correctness~\citep{ye2025justice,Saito2023VerbosityBI}. These concerns are particularly acute in domains requiring factual precision and safety.

In biomedical NLP, only a handful of studies have assessed LLMs as judges. Some works report encouraging alignment with human annotators on medical QA dimensions, while also noting substantial weaknesses in subjective or safety-critical categories such as empathy, harm assessment, and contextual clinical reasoning~\citep{diekmann-etal-2025-llms}. Other research highlights reduced agreement with experts on specialized knowledge tasks, reinforcing the need for human oversight~\citep{szymanski2024limitationsllmasajudgeapproachevaluating}. Additional efforts toward automated medical QA evaluation suggest that LLMs can reduce annotation burden but remain limited in complex or domain-specific cases~\citep{krolik2024leveraginglargelanguagemodels}.

To our knowledge, no prior work has investigated LLM-as-a-Judge for French medical OEQA. Our study complements existing work in the medical domain by focusing on a non-English language and by framing evaluation as a binary semantic equivalence judgment between a generated answer and a reference, rather than graded scoring or safety annotation, thereby broadening the scope of automated medical QA evaluation.

\subsection{Model Alignment for Evaluation}
Most LLM-as-a-judge systems rely on SFT on large GPT-4~\citep{achiam2023gpt} derived judgment datasets. Prometheus~\citep{kim2024prometheus} uses SFT on rubric-based feedback to induce evaluation capabilities comparable to GPT-4. JudgeLM~\cite{zhu2025judgelm} similarly applies SFT on GPT-4–generated judgments with bias-mitigation strategies, achieving strong agreement with its teacher judge. PandaLM~\citep{wang2024pandalm} also adopts an SFT approach, training a lightweight evaluator on human preference data to offer an efficient alternative.

While these systems illustrate the effectiveness of SFT for building evaluator models, reinforcement optimization remains largely unexplored for judgment tasks, especially in biomedical NLP where domain expertise and safety considerations are essential. Motivated by this gap, we apply GRPO~\citep{shao2024deepseekmath} to align a small LM (Phi-3.5-mini-instruct~\citep{abdin2024phi3technicalreporthighly}) on expert French medical OEQA annotations, exploring whether reinforcement-based alignment can complement SFT-only approaches.

\section{Experimental Setup}
\subsection{Task Definition}
\label{subsec:task-def}
We frame the evaluation of OEQA as a binary equivalence judgment: given a question, a ground-truth reference answer, and a candidate answer produced by an LLM, the evaluator must decide whether the candidate is semantically equivalent to the reference in terms of medical correctness and clinical validity. An example of the question, its ground truth answer, and the generated answers, are given in Appendix~\ref{app:equivalence_example}. We adopt this binary formulation for two reasons. First, it provides a practical and cognitively light annotation protocol for the medical expert, who can reliably determine equivalence without producing detailed rubrics or multi-point scores. This could make large-scale expert annotation feasible in a demanding domain. Second, binary equivalence offers a simple and interpretable target for LLM evaluators, avoiding the instability and ambiguity that often arise with graded scoring schemes. It also mirrors the behavior of exact match evaluation commonly used in multiple-choice QA: the reference answer is treated as the canonical correct solution, and the model output is judged according to whether it preserves the medically relevant meaning.

\subsection{Dataset}
Our study relies on two datasets: a small training set used for aligning the evaluator model, and a larger evaluation set used to benchmark LLM-as-a-Judge performance.

\paragraph{Training Set} The training set consists of 100 French OEQA instances scraped from S-EDITION\footnote{\url{https://s-editions.fr/}}, an online platform providing educational material for medical students. For each question, we generated a candidate answer using a medical-adapted version of Qwen3-4B-Instruct-2507~\footnote{\url{https://huggingface.co/MedInjection-FR/QWEN-4B-ALL}}. The clinician was then presented with the triplets (question, reference answer, generated answer) and annotated semantic equivalence following the protocol described in~\ref{subsec:task-def}.

To increase the diversity of training signals, we constructed 42 contrastive negative examples by swapping answers between originally positive (equivalent) and negative (non-equivalent) instances. We also created 42 additional positive examples by paraphrasing the reference answer with GPT-4o~\cite{achiam2023gpt} and treating the paraphrased version as the generated answer. This augmentation strategy expands the range of valid equivalence patterns while maintaining strict medical correctness. The paraphrasing prompt is provided in Appendix~\ref{app:para-prompt}.

\paragraph{Evaluation Set} The evaluation set contains 500 OEQA instances. We first extracted 100 questions and their reference answers from the MediQAl~\citep{bazoge2025mediqal} dataset. For each question, we then generated five answers, one from each of five LLMs: Gemma-3-4B-Instruct~\citep{gemmateam2025gemma3technicalreport}, a medically adapted version of LLaMA-13B-Chat (referred to as SFT-LLaMA-13B-Chat, not provided for anonymity), MedGemma-4B-Instruct~\citep{sellergren2025medgemma}, SFT-Qwen-4B-Instruct, and Qwen3-4B-Instruct-2507~\citep{qwen3technicalreport}. This procedure yields 5 responses per question, resulting in 500 model-generated answers that were annotated by a board-certified physician specialized in neurovascular medicine, with five years of senior clinical experience.

To assess annotation consistency, a subset of 10 items was independently annotated by a second clinician. The two annotators agreed on 9 out of 10 cases (\textbf{90\%} agreement), with a single disagreement.

\subsection{LLM Evaluators}
We evaluate a diverse set of LLMs as judges, covering both proprietary and open-source models, as well as general-purpose and medically adapted architectures. The evaluator models include GPT-5.1\footnote{\url{https://openai.com/index/gpt-5-1/}}, Gemini-2.5-Pro~\citep{comanici2025gemini}, Qwen3-Next-80B-A3B-Instruct~\citep{qwen3technicalreport}, Phi-3.5-mini-instruct~\citep{abdin2024phi3technicalreporthighly} and Medgemma-27b-text-it~\citep{sellergren2025medgemma}. This selection enables us to examine how factors such as parameter scale, domain specialization, and training data influence alignment with expert medical decisions. By comparing generalist models to medically oriented ones, we assess whether reliable factual judgment in French medical QA is driven primarily by model size or by medical pretraining.

\subsection{Prompting Strategies}

We use a direct equivalence prompt that is strictly aligned with the task definition and the human annotation protocol. The evaluator is provided with the question, the reference answer, and the generated answer, and is required to output 1 if the two answers are equivalent, 0 otherwise. Using the same formulation for both human and model evaluation ensures conceptual consistency between annotation and automated judging, allowing agreement scores to be interpreted directly. This prompt also minimizes instruction complexity, reducing variability introduced by prompt design.

For LLM evaluators, we provide an English version of the prompt since all evaluated models are primarily trained on English and their capabilities in French may be uneven or uncertain. This choice ensures a more controlled and comparable evaluation across models. The full prompt is provided in Appendix~\ref{app:eval-prompt}.

\subsection{Evaluation Metrics}
To assess the alignment between LLM judges and the medical expert, we use accuracy as the primary evaluation metric, capturing the overall proportion of correct equivalence decisions. We additionally report Pearson correlation between model predictions and human labels, providing a rank-based measure of agreement that is insensitive to class imbalance and reflects consistency in the relative ordering of judgments. Beyond quantitative metrics, we conduct a targeted error analysis by manually inspecting cases where evaluators disagree with the clinician. This qualitative analysis aims to uncover systematic failure modes, such as over-sensitivity to paraphrasing, unwarranted penalization of concise but clinically valid answers, or medical hallucinations introduced during the evaluation process.

\subsection{LLM Alignment}
Beyond evaluating existing LLM judges, we investigate whether a lightweight model trained on a small dataset (184 instances) can be turned into a reliable evaluator through reinforcement alignment. We first fine-tune Phi-3.5-mini-instruct with SFT for 5 epochs, stopping when the validation loss plateaued. We then apply GRPO for an additional 2 epochs to further align the model with expert equivalence annotations. Importantly, the adaptation data is kept strictly disjoint from the 500-instance evaluation set. This choice avoids adapting to the answer-generating models used at evaluation time, allowing us to assess whether alignment improves general judgment behavior rather than overfitting to generation biases.

The goal of this two-stage procedure (\emph{SFT+GRPO}) is to encourage the model not only to reproduce the expert’s binary decisions but also to learn more stable and discriminative decision boundaries than those obtained through SFT alone.

This experiment is an exploratory study of reinforcement-based alignment for evaluator models in French medical OEQA. Additional ablation experiments are reported in Appendix~\ref{app:alignment_experiments}, and hyperparameter details are provided in Appendix~\ref{app:appendix-hparams}.

\subsection{Statistical Significance Testing}
\label{sec:significance}

To assess whether GRPO alignment leads to statistically significant improvements, we conduct paired significance tests across three model comparisons: \textit{base} vs.\ \textit{SFT-only}, \textit{base} vs.\ \textit{GRPO-aligned}, and \textit{SFT-only} vs.\ \textit{GRPO-aligned}. All tests are performed in a paired setting on the test set, with each model evaluated on identical test instances.

For accuracy, we use the \textit{exact McNemar test}, which is appropriate for paired binary predictions. Let $b_{01}$ denote the number of instances where model~A is incorrect and model~B is correct, and $b_{10}$ the reverse. The two-sided exact p-value is computed from the binomial distribution over the $b_{01} + b_{10}$ discordant pairs.

For F1-score and correlation, we apply \textit{paired bootstrap} and \textit{paired permutation} tests, as the sampling distributions of these metrics are not analytically tractable. In the bootstrap test, we draw $B = 10{,}000$ samples with replacement from the test set and compute the metric difference
\[
\Delta = m(H, P_A) - m(H, P_B),
\]
where $H$ denotes human labels and $P_A$, $P_B$ the predictions of the two models. The p-value is estimated as
\[
p = \mathbb{P}(|\Delta^{(b)}| \ge |\Delta_{\text{obs}}|).
\]

In the permutation test, we generate $K = 10{,}000$ permutations by randomly swapping model predictions at the instance level and recomputing $\Delta$. This test evaluates the null hypothesis that both models are exchangeable with respect to the evaluation metric.

Together, these complementary tests allow us to verify whether observed gains from SFT and GRPO alignment are statistically reliable across accuracy, F1-score, and correlation with expert judgments.

\section{Results \& Discussion}
\label{sec:result}
 
\begin{table*}[!ht]
    \centering
    \begin{tabular}{lccccc}
    \hline
    \textbf{Judge} & \textbf{P} & \textbf{R} & \textbf{F1} & \textbf{Accuracy} & \textbf{Pearson $r$} \\
    \hline
    MedGemma-27B        & 52.17 & 72.00 & \textbf{60.50} & 71.80 & 40.67 \\
    Qwen-80B            & 64.62 & 56.00 & 60.00 & \textbf{77.60} & \textbf{44.77} \\
    Gemini-2.5-Pro      & 65.35 & 44.00 & 52.59 & 76.20 & 38.81 \\
    GPT-5.1             & \textbf{76.19} & 32.00 & 45.07 & 76.60 & 38.27 \\
    Phi-3.5-mini             & 35.94 & \textbf{98.00}\textbf{*} & 52.59 & 47.00 & 27.49 \\
   \hline
    SFT-Phi-3.5-mini         & 38.81 & 91.33 & 54.47 & 54.20 & 29.79 \\
    GRPO-Phi-3.5-mini        & 51.91\textbf{*} & 63.33 & 57.06\textbf{*} & 71.40\textbf{*} & 36.33\textbf{*} \\
    \hline
    \end{tabular}
    \caption{Agreement between LLM judges and the medical expert on binary equivalence judgments. Best scores are shown in bold. The symbol \textbf{*} indicates the best-performing variant among the base, SFT-aligned, and GRPO-aligned models.}
    \label{tab:llm_judges_main}
\end{table*}

We evaluate the ability of different LLMs to act as automatic judges for semantic equivalence in French medical OEQA. Table~\ref{tab:llm_judges_main} reports the agreement metrics between each LLM judge and the expert's annotations on the evaluation set of 500 instances.

\subsection{Limits of Similarity-Based Evaluation}

\begin{table}[t]
\centering
\begin{tabular}{lc}
\hline
\textbf{Metric} & \textbf{Pearson $r$} \\
\hline
ROUGE-L        & 25.40 \\
BLEU           & 17.16 \\
BERTScore-F1   & 14.88 \\
\hline
\end{tabular}
\caption{Pearson correlation between automatic similarity metrics and expert equivalence judgments.}
\label{tab:metric_pearson_percentage}
\end{table}

To evaluate whether standard automatic metrics align with expert annotations of answer equivalence, we compute Pearson correlations between each metric score and binary equivalence labels assigned by medical experts, and report the results in Table~\ref{tab:metric_pearson_percentage}. ROUGE-L, BLEU and BERTScore's F1 exhibit weak correlations with expert judgments, indicating that lexical overlap and embedding-based similarity are poor indicators of clinical equivalence. In the medical QA setting, clinically equivalent answers may differ substantially in form, while surface-level similarity can mask missing or inaccurate clinical information. Consequently, automatic metrics fail to differentiate equivalent from non-equivalent answers according to expert criteria.

\subsection{Bias Analysis Across Generating Models}

\begin{figure}[t]
    \centering
    \includegraphics[width=\linewidth]{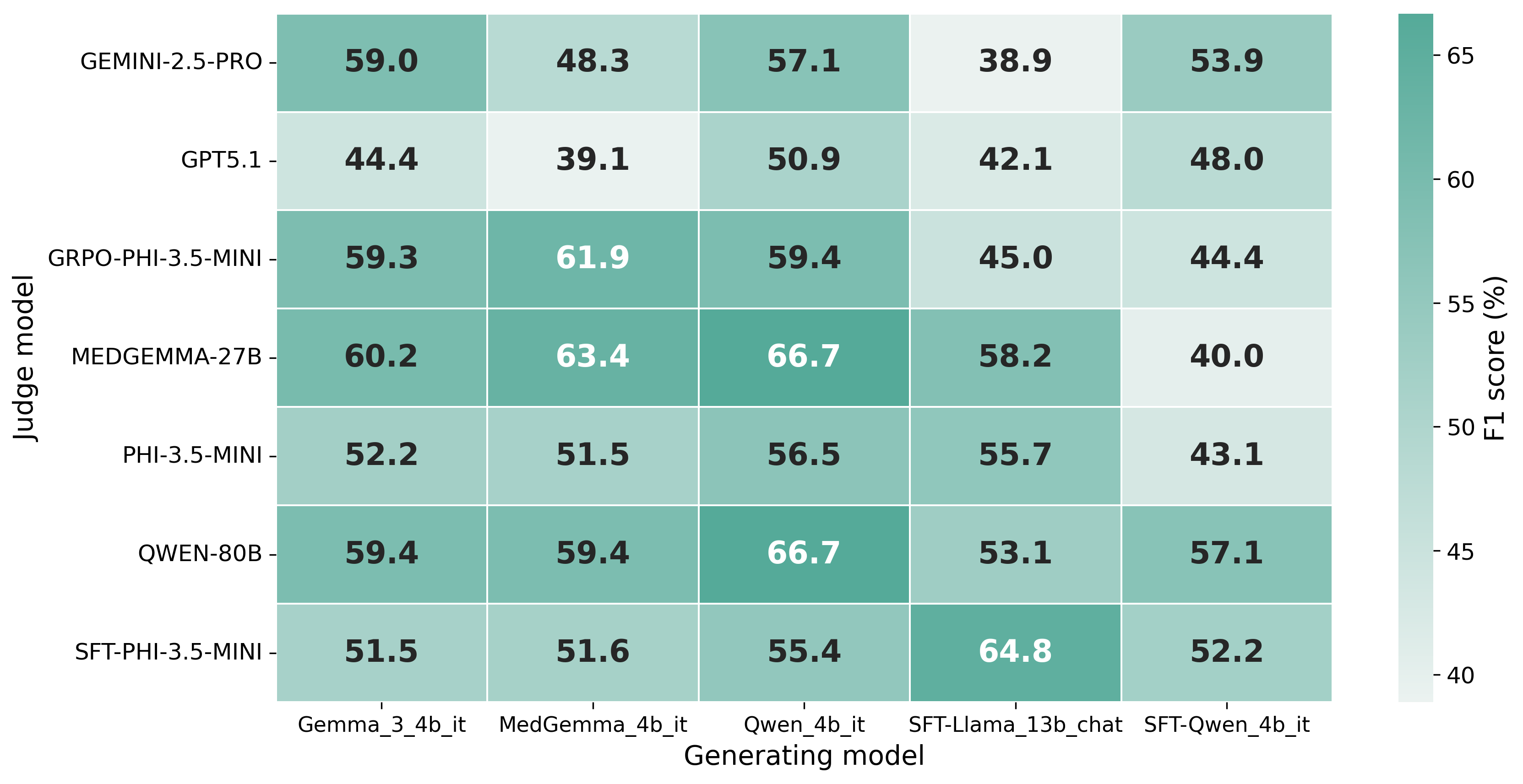}
    \caption{Heatmap of F1 scores for each judge model across answer-generating models.}
    \label{fig:judge_generator_heatmap}
\end{figure}

Since each question is associated with multiple answers generated by different answer-generating models, we analyze LLM judges at a finer granularity by reporting their performance separately for each generating model. This allows us to identify potential biases toward specific answer styles or model families. Results are shown in the heatmap of Figure~\ref{fig:judge_generator_heatmap}, and detailed results are reported in Appendix~\ref{app:judge-generator}.

Overall, none of the evaluated judges is fully invariant to the source of answer generation. All models show noticeable variation in precision, recall, and F1 score across different generators, indicating that LLM-based evaluation is influenced not only by semantic content but also by characteristics of the generated answers.

Closed-access models such as GPT-5.1 and Gemini-2.5-Pro exhibit relatively stable accuracy across generators, but their precision–recall balance varies. In particular, both models achieve high precision but lower recall on answers produced by fine-tuned Llama models, suggesting a conservative tendency to reject equivalence for more concise generations. Indeed, the fine-tuned models (Llama and Qwen) are less verbose compared to the remaining answers generators. A table showing the average mean of characters produced by each of the models is shown in Appendix~\ref{app:generator_len}.

MedGemma-27B shows a more consistent performance across generators, maintaining relatively stable F1 scores for both general-purpose and domain-adapted answers. Notably, it does not show a systematic preference for outputs from its own model family, indicating reduced sensitivity to surface-level variation. In contrast, Qwen-80B performs best on answers generated by Qwen-based models and shows lower agreement on outputs from other families, suggesting generator dependent bias.

Finally, the base Phi-3.5-mini model exhibits a strong bias toward predicting equivalence, with very high recall but low precision. Its F1 score varies across generators, with better performance on answers from fine-tuned Qwen and LLama models, reflecting sensitivity to differences specific to generators.

Overall, the analysis shows that LLM-based judges are not fully invariant to the source of generation, with noticeable variations in agreement across different answer-generating models

\subsection{LLM-as-a-Judge Evaluation}
\paragraph{Alignment with human judgments:} Across all models, a substantial variability is observed in alignment with the manual annotations. The best performing evaluators in terms of balanced agreement are MedGemma-27B and Qwen-80B, achieving the highest F1 scores (60.5\% and 60\% respectively) while maintaining a reasonable accuracy. These models also obtain the highest Pearson correlations, which suggests that they best capture the relative structure of expert judgment beyond simple label matching. Interestingly, MedGemma-27B is substantially smaller than Qwen-80B, yet it has been specifically adapted to the biomedical domain, whereas Qwen-80B is a larger, general-purpose model without such specialization.

In contrast, GPT-5.1, despite its remarkable performance on a diversity of tasks, is surpassed by both MedGemma and Qwen in balanced agreement with the expert annotations. While GPT-5.1 achieves high precision (76.2\%), its recall is very low (32.0\%), reflecting a conservative tendency to reject equivalence. This suggests that even state-of-the-art general-purpose models may struggle in domain-specific evaluation. In the same scope of closed-access models, Gemini-2.5-Pro, a thinking LLM, achieves a more balanced performance resulting in an F1 score of 52.6\%, and an accuracy of 76.2\%. Shifting to the smallest open-source model in the evaluation, Phi-3.5-mini, with only 3.8B parameters, achieves competitive performance. The model exhibits very high recall (98.0\%) but the lowest precision, yielding an F1 score of 52.6\% and an accuracy of 47.0\%. This is primarily caused by its tendency to over-predict the positive class (409 positives), which contrasts with the behavior observed in the other models.

\paragraph{Correlation with expert judgments:} Pearson correlation provides a complementary view of alignment that is less sensitive to class imbalance. Once again, Qwen and MedGemma achieve the highest correlations, indicating that they not only match individual labels more accurately but also better preserve the relative ordering of equivalence decisions across instances. Models with extreme precision/recall imbalance exhibit notably lower correlations, suggesting that biased decision strategies negatively impact global alignment with expert reasoning.

\subsection{Lightweight adaptation strategies}
\paragraph{Overall Results:} We selected Phi-3.5 for adaptation experiments for two main reasons. First, its compact size (3.8B parameters) makes it suitable for low-resource settings, where the available training data does not exceed 100 instances. Second, its base behavior differs markedly from the other models, exhibiting extremely high recall and low precision due to over-predicting the positive class (409 positives). The goal of these experiments is to assess whether lightweight fine-tuning can significantly improve performance under data scarcity.

Applying SFT alone yields a moderate improvement: the model partially corrects its tendency to overgenerate positive predictions, improving the F1 score to 54.47\% (Table~\ref{tab:llm_judges_main}). However, the most substantial gains are obtained after the additional GRPO adaptation step. This procedure further balances precision and recall, correcting the over-prediction of positives, bringing its overall performance above GPT and Gemini, and achieving an F1 score close to that of MedGemma. These results highlight that even extremely lightweight LLMs can achieve competitive evaluation performance when adapted to domain-specific tasks in low-resource scenarios.
\begin{figure}[t]
    \centering
    \includegraphics[width=\linewidth]{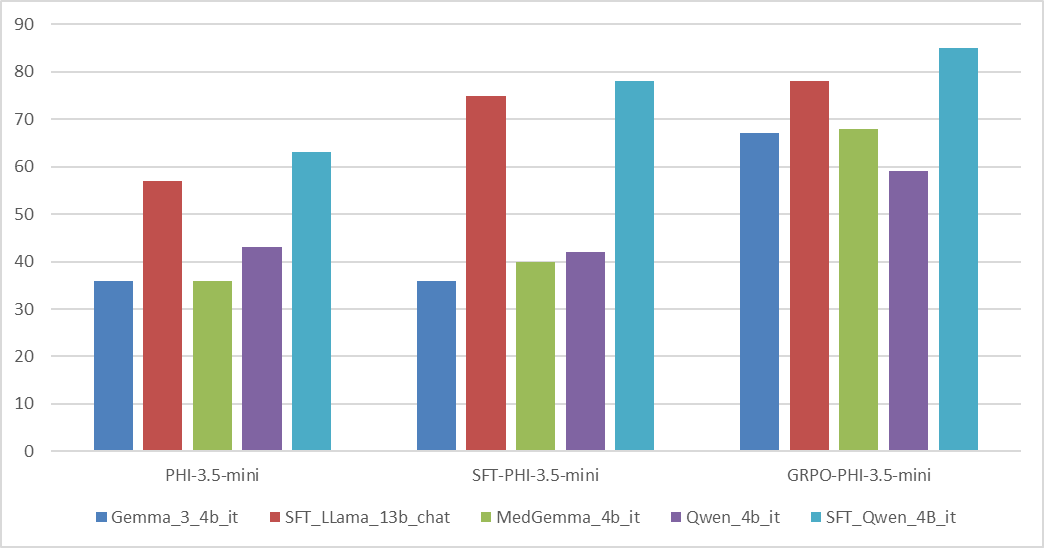}
    \caption{Comparison of F1 scores of Phi models on the text generated by multiple LLMs}
    \label{fig:phi-vs-generators}
\end{figure}

\paragraph{Impact of adaptation strategies on generating models:} To better understand how SFT and GRPO affect Phi-3.5-mini, we analyze its predictions on instances generated by each of the 5 answer generators. Results are shown in Figure~\ref{fig:phi-vs-generators}. Overall, the predictions are better on the less verbose models, the fine-tuned Llama and Qwen. Before any adaptation, Phi’s equivalence predictions were roughly similar for instances generated by Gemma-3-4B and MedGemma-4B, while instances generated by the base Qwen-4B achieved slightly higher equivalence. The finetuned Llama-13B instances surpassed the later, and the highest performance was observed on instances generated by the finetuned Qwen-4B model.

Applying SFT had the largest effect on instances generated by the finetuned Llama and finetuned Qwen models, partially correcting Phi’s over-prediction of positives. With the GRPO step, further light improvements are observed on these same instances, but the most significant gains occur for predictions on the remaining generations, particularly those from the Gemma family, bringing their equivalence predictions closer to those of the finetuned Qwen instances. Overall, these results indicate that lightweight adaptation not only improves Phi’s balance between precision and recall but also reduces variability across different types of generated instances, with the greatest impact on domain-relevant or underperforming generations.

\paragraph{Significance testing:} As shown in Table~\ref{tab:mcnemar}, McNemar’s exact test indicates a significant reduction in classification errors for both SFT and GRPO compared to the base model, with GRPO yielding the largest improvement. The difference between SFT and GRPO is also significant, confirming that GRPO provides additional corrective effects beyond supervised fine-tuning. However, bootstrap and permutation tests on F1 and correlation (Tables~\ref{tab:F1-significance} and~\ref{tab:r-significance}) do not show statistically significant differences ($p > 0.1$), suggesting that the adapted judges behave similarly in terms of positive-class performance and overall correlation with human annotations. Taken together, these results indicate that GRPO-Phi-3.5-mini is clearly superior in terms of overall classification accuracy, but its improvements do not translate into substantially better handling of the positive (equivalent) class.

\begin{table}[H]
\centering
\setlength{\extrarowheight}{1pt} 
\resizebox{\columnwidth}{!}{%
\begin{tabular}{ccccc}
\hline
\textbf{\begin{tabular}[c]{@{}c@{}}Comparison\\ (A vs. B)\end{tabular}} & \textbf{Acc (A)} & \textbf{Acc (B)} & \textbf{\begin{tabular}[c]{@{}c@{}}$\Delta$Acc\\ (Acc (B) - (Acc (A))\end{tabular}} & \textbf{McNemar p} \\ \hline
Base vs. SFT & 47.00 & 54.20 & 7.20 & 1.25E-06 \\ \hline
Base vs. GRPO & 47.00 & 71.40 & 24.40 & 1.44E-16 \\ \hline
SFT vs. GRPO & 54.20 & 71.40 & 17.20 & 2.64E-11 \\ \hline
\end{tabular}%
}
\caption{McNemar test results for accuracy differences between the base, SFT-only, and GRPO-aligned evaluators on the test set.}
\label{tab:mcnemar}
\end{table}

In conclusion, the results show that LLM judges are sensitive to the source of generated answers, with substantial variation in precision, recall, and F1 scores across different generators. Closed-access and general-purpose models such as GPT-5.1 and Gemini-2.5-Pro maintain stable accuracy but exhibit imbalanced decision behavior, often over-rejecting equivalence for certain answer styles. Domain-adapted models like MedGemma-27B achieve more consistent performance and reduced generator-dependent bias, while Qwen-80B performs best on answers from its own family, highlighting some preferential behavior. Smaller models, exemplified by Phi-3.5-mini, initially over-predict equivalence, but lightweight adaptation via SFT and GRPO significantly improves the balance between precision and recall, and reduces variability across generators. These findings underscore that both model size and adaptation strategy impact reliability as automatic judges, with carefully adapted models achieving performance and alignment comparable to larger or domain-specialized LLMs.

\section{Conclusion}
This work presents a preliminary investigation into the use of LLMs as judges for evaluating semantic equivalence in French medical OEQA. Our results demonstrate that current LLM judges vary widely in their alignment with expert annotations and are strongly influenced by the model that generated the answer being evaluated. Domain-adapted models such as MedGemma-27B achieve more consistent agreement with expert judgments than general-purpose models, which exhibit stronger generator-dependent biases. These results raise the question of whether current LLM judges are sufficiently accurate to be trusted beyond auxiliary evaluation roles.
We further showed that compact models can be transformed into effective evaluators with minimal supervision. Lightweight SFT and GRPO substantially improved the performance of Phi-3.5-mini, reducing variability across generators and achieving agreement levels comparable to much larger domain-adapted models. Significance testing confirmed that GRPO provides statistically meaningful gains in overall correctness, though improvements remain limited for positive-class detection.

While our dataset is small, this study provides evidence that generator-aware evaluation and alignment strategies are essential components for building reliable medical LLM judges in low-resource settings. Future work will scale the dataset, extend the evaluation to richer clinical tasks, explore multilingual generalization, and investigate more advanced optimization methods for robust, domain-specialized evaluators.

\section*{Limitations}
Although our findings provide initial insights into the feasibility of LLM-as-a-Judge for French medical OEQA, several limitations must be acknowledged.

First, the study relies on a relatively small number of expert annotations. While sufficient to reveal systematic trends, such as generator-dependent biases and the impact of lightweight adaptation,   the dataset is not large enough to capture the full variability of medical reasoning or to provide stable estimates of performance across all metrics. This also limits the statistical power of tests based on F1 score and correlation.

Second, our evaluation focuses exclusively on binary semantic equivalence. This simplifies the annotation process but does not capture more nuanced errors related to incompleteness, misleading statements, style, or clinical safety. A richer annotation schema would allow a more comprehensive assessment of LLM evaluators in real clinical QA settings.

Third, while our analysis highlights that Qwen-based judges exhibit a bias toward the Qwen generator family, we do not conduct a comparable bias analysis for other judges such as Phi-3.5 or GPT-5.1, as these models are used exclusively as evaluators and not as generators. Consequently, the absence of observed bias for these judges should not be interpreted as evidence of neutrality, but rather as an untested assumption, which may lead to overly optimistic expectations regarding their robustness as evaluators.

Fourth, we do not evaluate the judges in multilingual or cross-domain contexts. The conclusions drawn here may not generalize beyond French medical OEQA, particularly given the strong generator-dependence observed. Larger and more diverse datasets are needed to understand how widely these findings extend.

Finally, the alignment experiments are limited to one compact model (Phi-3.5-mini) and use only SFT and GRPO. Other preference optimization methods, additional reward formulations, or more diverse supervision signals may yield further improvements.

Despite these limitations, the study offers actionable insights and establishes a foundation for more extensive future work in scalable medical OEQA evaluation.

\section*{Ethical Considerations}

This work focuses on the evaluation of medical question answering systems and does not involve deploying LLMs for clinical decision-making. All medical judgments used in this study were provided by qualified clinicians acting in a research capacity, and the annotations reflect assessments of semantic equivalence rather than medical advice.

The datasets used consist of publicly available educational and benchmark resources, and no personally identifiable patient information was accessed or generated. The annotation protocol was designed to minimize cognitive load and risk for clinicians by using a binary equivalence task, and annotation was conducted independently of any downstream clinical application.

We emphasize that the LLM-based evaluators studied here exhibit only moderate agreement with expert judgments and are not suitable for use as autonomous evaluators in clinical or safety-critical settings. Their role is limited to supporting large-scale or exploratory evaluation, where expert review remains indispensable. Over-reliance on automated judges in medical contexts could obscure clinically meaningful errors, particularly given the generator-dependent biases identified in our analysis.

Finally, while our study explores lightweight adaptation of evaluator models, we intentionally keep training and evaluation data disjoint to reduce risks of overfitting to annotator or generator biases. All results are reported transparently, and code and data are released to encourage reproducibility and scrutiny. We view this work as a preliminary step toward more robust and ethically grounded evaluation frameworks for medical LLMs.

\section*{Acknowledgments}
This work was financially supported by ANR
MALADES (ANR-23-IAS1-0005). It was provided with computing HPC and storage resources by GENCI at IDRIS thanks to the grants 2025-AD011015256R1 and 2025-AD011016540 on the supercomputer Jean Zay's H100 partition.
 \newpage

\bibliography{custom}
\newpage
\appendix

\section{Training Hyperparameters}
\label{app:appendix-hparams}

Table~\ref{tab:hparams} summarizes the hyperparameters used for SFT and GRPO alignment of the \textsc{Phi-3.5-mini-instruct} evaluator.

\begin{table}[h]
\centering
\small
\begin{tabular}{lcc}
\hline
\textbf{Hyperparameter} & \textbf{SFT} & \textbf{GRPO} \\
\hline
Training set size & 166 & 184 \\
Validation set size & 18 & -- \\
Number of epochs & 5 & 2 \\
Learning rate & $5 \times 10^{-6}$ & $1 \times 10^{-5}$ \\
Batch size & 1 & 1 \\
Gradient accumulation steps & 4 & 4 \\
Max sequence / prompt length & 1024 & 1024 \\
Max completion length & -- & 8 \\
Learning rate scheduler & Cosine & -- \\
Warmup ratio & 0.05 & -- \\
Number of generations & -- & 4 \\
\hline
\end{tabular}
\caption{Hyperparameters used for SFT and GRPO alignment of the \textsc{Phi-3.5} evaluator.}
\label{tab:hparams}
\end{table}

\subsection{GRPO Reward Function}

The GRPO reward combines prediction correctness and output format constraints:
\[
R = R_{\text{correctness}} + R_{\text{format}},
\]
where $R_{\text{correctness}} = +1.0$ if the predicted label matches the gold label and $-1.0$ otherwise, and $R_{\text{format}} = +0.5$ if the output is exactly \texttt{0} or \texttt{1}, and $0$ otherwise.

\section{Prompt Templates}
\label{sec:appendix-prompts}

We report the prompt templates used for data augmentation and automated evaluation.

\newtcolorbox{promptbox}[1]{
  colback=gray!7,        
  colframe=gray!90,      
  boxrule=0.6pt,
  arc=3pt,
  left=6pt,
  right=6pt,
  top=6pt,
  bottom=6pt,
  title=\textbf{#1},
  fonttitle=\small,
}

\subsection{Paraphrasing Prompt}
\label{app:para-prompt}
\begin{promptbox}{System prompt (paraphrasing)}
\begin{lstlisting}
Tu es un assistant médical francophone.
Ta tâche est de PARAPHRASER la réponse de référence sans en changer le sens médical.
Ne modifie pas les informations clés.
Ne réponds qu’avec la version paraphrasée.
\end{lstlisting}
\end{promptbox}

\subsection{Evaluation Prompt}
\label{app:eval-prompt}
\begin{promptbox}{System prompt (evaluation)}
\begin{lstlisting}
You are a medical evaluator tasked with assessing whether a candidate answer is equivalent to the ground truth.
Assign a score strictly according to the criteria below. Do not include any explanations, comments, or extra text in your response.

Scoring criteria:
Score 0: Not equivalent
Score 1: Equivalent

Definition of equivalence:
Two answers are considered equivalent if the essential expected information is covered. Minor differences in wording, or additional or missing details, are acceptable as long as the candidate answer would be considered an acceptable response to the question.

Return only the score (0 or 1), nothing else.
\end{lstlisting}
\end{promptbox}

\section{Significance Testing Results}
\begin{table}[H]
\centering
\setlength{\extrarowheight}{1pt} 
\resizebox{\columnwidth}{!}{%
\begin{tabular}{cccccc}
\hline
\textbf{\begin{tabular}[c]{@{}c@{}}Comparison\\ (A vs. B)\end{tabular}} & \textbf{Pearson r(A)} & \textbf{Pearson r(B)} & \textbf{\begin{tabular}[c]{@{}c@{}}$\Delta$Prearson r\\ (r(B) - r(A))\end{tabular}} & \textbf{Bootstrap  p} & \textbf{Permutation p} \\ \hline
Base vs. SFT & 27.49 & 29.79 & 2.3 & 0.5533 & 0.4255 \\ \hline
Base vs. GRPO & 27.49 & 36.33 & 8.84 & 0.4985 & 0.1106 \\ \hline
SFT vs. GRPO & 29.79 & 36.33 & 6.54 & 0.5138 & 0.1899 \\ \hline
\end{tabular}%
}
\caption{Paired bootstrap and permutation test results for correlation differences between the base, SFT-only, and GRPO-aligned evaluators on the test set.}
\label{tab:r-significance}
\end{table}
\begin{table}[h]
\centering
\setlength{\extrarowheight}{1pt} 
\resizebox{\columnwidth}{!}{%
\begin{tabular}{cccccc}
\hline
\textbf{\begin{tabular}[c]{@{}c@{}}Comparison\\ (A vs. B)\end{tabular}} & \textbf{F1 (A)} & \textbf{F1 (B)} & \textbf{\begin{tabular}[c]{@{}c@{}}$\Delta$F1\\ (F1 (B) - F1 (A))\end{tabular}} & \textbf{Bootstrap  p} & \textbf{Permutation p} \\ \hline
Base vs. SFT & 52.59 & 54.47 & 1.88 & 0.5049 & 0.092 \\ \hline
Base vs. GRPO & 52.59 & 57.06 & 4.47 & 0.5002 & 0.1278 \\ \hline
SFT vs. GRPO & 54.47 & 57.06 & 2.59 & 0.5338 & 0.3389 \\ \hline
\end{tabular}%
}
\caption{Paired bootstrap and permutation test results for F1-score differences between the base, SFT-only, and GRPO-aligned evaluators on the test set.}
\label{tab:F1-significance}
\end{table}

\section{Additional Alignment Experiments}
\label{app:alignment_experiments}

To better understand the effect of different alignment schedules, we conducted several exploratory experiments combining SFT and GRPO on Phi-3.5-mini. These experiments were designed to assess whether the duration of SFT influence the effectiveness of subsequent reinforcement-based alignment.

We considered three alignment variants: (i) a short SFT phase followed by GRPO, (ii) GRPO applied directly without any prior SFT, and (iii) extended SFT until the validation loss stabilized, followed by GRPO. Table~\ref{tab:alignment_ablation} summarizes the aggregated confusion statistics and F1 scores for each configuration.
\begin{table*}[h]
\centering
\setlength{\extrarowheight}{1pt} 
\small
\begin{tabular}{lccccc}
\hline
\textbf{Model Variant} & TP & FP & TN & FN & \textbf{F1} \\
\hline
PHI-3.5 (Base) & 147 & 262 & 88 & 3 & 0.526 \\

PHI-3.5 + GRPO (after SFT-60 steps) & 142 & 246 & 104 & 8 & 0.528 \\

PHI-3.5 + GRPO (after SFT to convergence) & 95 & 88 & 262 & 55 & \textbf{0.571} \\

PHI-3.5 + GRPO (from scratch) & 139 & 219 & 131 & 11 & 0.547 \\
\hline
\end{tabular}

\caption{Ablation of alignment schedules for Phi-3.5-mini. We compare the base model, GRPO applied from scratch, GRPO after a short SFT warm-up, and GRPO applied after SFT convergence. The strongest performance is obtained when GRPO is applied after SFT loss stabilization.}
\label{tab:alignment_ablation}
\end{table*}

Across experiments, we observe that applying GRPO without sufficient prior SFT leads to unstable behavior. Although direct GRPO from scratch improves recall relative to the base model, it does not consistently balance precision and recall, resulting in lower overall F1. A short SFT warm-up (60 steps) before GRPO yields only marginal improvements, suggesting that limited exposure is insufficient to meaningfully shape the model’s decision boundaries.

The best performance is obtained when GRPO is applied \emph{after} SFT has converged and the validation loss has stabilized. In this setting, SFT appears to provide a strong initialization that clarifies the task structure and reduces noise in the reward signal, allowing GRPO to refine decision boundaries more effectively. This finding is consistent with prior work on reinforcement learning for language models, which commonly recommends an initial SFT phase to ensure instruction-following and output format stability before applying policy optimization~\citep{shao2024deepseekmath,kang-etal-2025-grpo}.

It is worth noting that Phi-3.5-mini already follows the required binary output format in its base form, meaning that SFT is not strictly necessary for format compliance in our task. Instead, we hypothesize that the main role of SFT here is to reduce variance in early GRPO updates and anchor the policy closer to expert-labeled decision regions. Once this stabilization is achieved, GRPO can more reliably correct systematic biases, such as over-prediction of equivalence, leading to the strongest overall performance.

These exploratory results motivate our final alignment strategy, which applies GRPO only after SFT convergence. While preliminary, they suggest that even in low-resource settings, the ordering and duration of alignment stages play a critical role in achieving stable and reliable evaluator behavior.

\section{Illustrative Example of Semantic Equivalence}
\label{app:equivalence_example}

Table~\ref{tab:equivalence_example} presents an illustrative example from the evaluation set, highlighting the challenges of assessing semantic equivalence in French medical OEQA. The example shows multiple model-generated answers to the same question, all judged against a single ground-truth reference by a clinician. Despite substantial variation in formulation, verbosity, and technical framing, several answers are annotated as equivalent, illustrating that equivalence depends on the presence of essential medical information rather than surface form.

\begin{table*}[t]
\centering
\footnotesize
\begin{tabular}{p{0.97\textwidth}}
\hline
\textbf{Question} \\
\emph{Citez par quels moyens réaliser une laryngoscopie indirecte ?} \\[0.3em]

\textbf{Clinical Context} \\
Un homme de 60 ans consulte pour une dysphagie aux solides s’aggravant depuis 2 mois. Il a perdu 6 kg en 6 mois. L’examen ORL par laryngoscopie indirecte retrouve une lésion d’allure tumorale du sinus piriforme droit. \\[0.6em]

\textbf{Ground Truth (Expert Reference)} \\
-- Au miroir (miroir laryngé avec lumière frontale, de Clar) \\
-- Au nasofibroscope \\[0.6em]

\textbf{Model Answers and Expert Judgments} \\[0.3em]

\textbf{Gemma-3-4B-Instruct:} \\
\emph{En se basant sur l'instruction médicale demandée ("Citez par quels moyens réaliser une laryngoscopie indirecte ?") et le contexte fourni (...)} \\
La laryngoscopie indirecte, également appelée laryngoscopie par la voix ou laryngoscopie phonodynamique, est réalisée en utilisant la phonation du patient. Elle consiste à : \\
1. Poser le patient en position assise... \\
2. Placer un otoscope... \\
3. Demander au patient de produire des sons vocaux... \\
4. Observer attentivement les structures laryngées... \\
\textbf{Important :} La laryngoscopie indirecte est un outil diagnostique utile, mais elle offre une visualisation limitée. \\
\emph{Expert equivalence: Yes} \\[0.4em]

\emph{Summary: Long procedural description; essential means are indirectly covered.} \\[0.6em]

\textbf{SFT-LLaMA-13B-Chat:} \\
La laryngoscopie indirecte peut être réalisée par endoscopie ou par examen visuel à l'aide d'un laryngoscope. \\
\emph{Expert equivalence: Yes} \\[0.4em]

\emph{Summary: Very concise answer mentioning endoscopic and visual means.} \\[0.6em]

\textbf{MedGemma-4B-Instruct:} \\
Pour réaliser une laryngoscopie indirecte, les étapes suivantes sont nécessaires : \\
1. Préparation... \\
2. Technique... \\
3. Documentation... \\
\textbf{Important:} Cette réponse est basée sur l'instruction médicale fournie et ne remplace pas un avis médical professionnel. \\
\emph{Expert equivalence: Yes} \\[0.4em]

\emph{Summary: Structured procedural explanation; means are implied rather than explicitly listed.} \\[0.6em]

\textbf{Qwen-4B-Instruct:} \\
La laryngoscopie indirecte est réalisée en utilisant un miroir et une lumière... \\
\textbf{En résumé :} \\
-- L’utilisation d’un miroir de laryngoscopie indirecte \\
-- Une source de lumière \\
\emph{Expert equivalence: Yes} \\[0.4em]

\emph{Summary: Explicit mention of the expected instruments.} \\[0.6em]

\textbf{SFT-Qwen-4B-Instruct:} \\
L'irrigation de la langue, du pharynx et du larynx \\
\emph{Expert equivalence: No} \\[0.4em]

\emph{Summary: Incorrect procedure; does not correspond to indirect laryngoscopy.} \\

\hline
\end{tabular}
\caption{Illustrative example of semantic equivalence judgments in French medical OEQA. Answers with widely different formulations, levels of detail, and terminology can still be judged equivalent by the clinician as long as the essential medical content is preserved, while concise but incorrect answers are rejected.}
\label{tab:equivalence_example}
\end{table*}

This example illustrates why surface-level automatic metrics are insufficient for medical OEQA evaluation. Answers that differ substantially in wording or verbosity may remain clinically valid, whereas superficially similar responses can fail to convey the required medical meaning.

\section{Detailed Results for Judge Bias Analysis}
\subsection{Overall Results}
\label{app:judge-generator}
\begin{table*}[t]
\centering
\setlength{\extrarowheight}{1pt} 
\small
\begin{tabular}{llccccc}
\hline
\textbf{Judge} & \textbf{Generator} & Accuracy & P & R & F1 & Spearman $\rho$ \\
\hline
GEMINI-2.5-PRO & Gemma-3-4B-IT            & 75.00 & 69.23 & 51.43 & 59.02 & 42.54 \\
GEMINI-2.5-PRO & SFT-LLaMA-13B-Chat       & 78.00 & 100.00 & 24.14 & 38.89 & 42.93 \\
GEMINI-2.5-PRO & MedGemma-4B-IT           & 70.00 & 58.33 & 41.18 & 48.28 & 28.87 \\
GEMINI-2.5-PRO & Qwen-4B-IT               & 70.00 & 60.61 & 54.05 & 57.14 & 34.31 \\
GEMINI-2.5-PRO & SFT-Qwen-4B              & 88.00 & 63.64 & 46.67 & 53.85 & 47.89 \\
\hline
GPT-5.1        & Gemma-3-4B-IT            & 75.00 & 100.00 & 28.57 & 44.44 & 45.43 \\
GPT-5.1        & SFT-LLaMA-13B-Chat       & 78.00 & 88.89 & 27.59 & 42.11 & 41.51 \\
GPT-5.1        & MedGemma-4B-IT           & 72.00 & 75.00 & 26.47 & 39.13 & 31.96 \\
GPT-5.1        & Qwen-4B-IT               & 71.00 & 68.18 & 40.54 & 50.85 & 34.30 \\
GPT-5.1        & SFT-Qwen-4B              & 87.00 & 60.00 & 40.00 & 48.00 & 42.01 \\
\hline
MEDGEMMA-27B   & Gemma-3-4B-IT            & 67.00 & 52.08 & 71.43 & 60.24 & 34.41 \\
MEDGEMMA-27B   & SFT-LLaMA-13B-Chat       & 77.00 & 61.54 & 55.17 & 58.18 & 42.50 \\
MEDGEMMA-27B   & MedGemma-4B-IT           & 70.00 & 54.17 & 76.47 & 63.41 & 40.90 \\
MEDGEMMA-27B   & Qwen-4B-IT               & 66.00 & 52.31 & 91.89 & 66.67 & 43.21 \\
MEDGEMMA-27B   & SFT-Qwen-4B              & 79.00 & 35.00 & 46.67 & 40.00 & 28.01 \\
\hline
PHI-3.5-MINI   & Gemma-3-4B-IT            & 36.00 & 35.35 & 100.00 & 52.24 & 7.37 \\
PHI-3.5-MINI   & SFT-LLaMA-13B-Chat       & 57.00 & 39.71 & 93.10 & 55.67 & 34.39 \\
PHI-3.5-MINI   & MedGemma-4B-IT           & 36.00 & 34.69 & 100.00 & 51.52 & 10.25 \\
PHI-3.5-MINI   & Qwen-4B-IT               & 43.00 & 39.36 & 100.00 & 56.49 & 19.36 \\
PHI-3.5-MINI   & SFT-Qwen-4B              & 63.00 & 28.00 & 93.33 & 43.08 & 36.41 \\
\hline
SFT-PHI-3.5-MINI & Gemma-3-4B-IT          & 36.00 & 35.05 & 97.14 & 51.52 & 0.61 \\
SFT-PHI-3.5-MINI & SFT-LLaMA-13B-Chat     & 75.00 & 54.76 & 79.31 & 64.79 & 48.31 \\
SFT-PHI-3.5-MINI & MedGemma-4B-IT         & 40.00 & 35.56 & 94.12 & 51.61 & 9.85 \\
SFT-PHI-3.5-MINI & Qwen-4B-IT             & 42.00 & 38.71 & 97.30 & 55.38 & 12.91 \\
SFT-PHI-3.5-MINI & SFT-Qwen-4B            & 78.00 & 38.71 & 80.00 & 52.17 & 44.51 \\
\hline
GRPO-PHI-3.5-MINI & Gemma-3-4B-IT         & 67.00 & 52.17 & 68.57 & 59.26 & 33.23 \\
GRPO-PHI-3.5-MINI & SFT-LLaMA-13B-Chat    & 78.00 & 81.82 & 31.03 & 45.00 & 40.92 \\
GRPO-PHI-3.5-MINI & MedGemma-4B-IT        & 68.00 & 52.00 & 76.47 & 61.90 & 38.00 \\
GRPO-PHI-3.5-MINI & Qwen-4B-IT            & 59.00 & 46.88 & 81.08 & 59.41 & 27.27 \\
GRPO-PHI-3.5-MINI & SFT-Qwen-4B           & 85.00 & 50.00 & 40.00 & 44.44 & 36.20 \\
\hline
QWEN-80B       & Gemma-3-4B-IT            & 74.00 & 65.52 & 54.29 & 59.38 & 40.89 \\
QWEN-80B       & SFT-LLaMA-13B-Chat       & 77.00 & 65.00 & 44.83 & 53.06 & 39.67 \\
QWEN-80B       & MedGemma-4B-IT           & 74.00 & 63.33 & 55.88 & 59.38 & 40.54 \\
QWEN-80B       & Qwen-4B-IT               & 75.00 & 65.79 & 67.57 & 66.67 & 46.68 \\
QWEN-80B       & SFT-Qwen-4B              & 88.00 & 61.54 & 53.33 & 57.14 & 50.38 \\
\hline
\end{tabular}
\caption{Detailed agreement metrics between LLM judges and the expert annotations, broken down by answer-generating model.}
\label{tab:judge_generator_results}
\end{table*}

Table~\ref{tab:judge_generator_results} provides the detailed agreement metrics between LLM judges and the expert annotations.
\subsection{Verbosity of Answers Generators}
\label{app:generator_len}
\begin{table}[H]
\centering
\setlength{\extrarowheight}{1pt} 
\small
\begin{tabular}{lc}
\hline
\textbf{Generating Model} & \textbf{Mean \# Characters} \\
\hline
Gemma-3-4B-IT & 1852.88 \\
SFT-LLaMA-13B-Chat & 121.26 \\
MedGemma-4B-IT & 1909.31 \\
Qwen-4B-IT & 1620.27 \\
SFT-Qwen-4B & 296.27 \\
\hline
\end{tabular}
\caption{Average number of characters per answer for each generating model.}
\label{tab:generator_length}
\end{table}

\end{document}